\documentclass[11pt,a4paper]{article}

\usepackage[T1]{fontenc}
\usepackage[utf8]{inputenc}

\usepackage{amsmath,amssymb,amsthm}
\usepackage{graphicx}
\usepackage{booktabs}
\usepackage{algorithm}
\usepackage{algpseudocode}
\usepackage{multirow}
\usepackage{caption}
\usepackage{subcaption}
\usepackage{textgreek}
\usepackage{url}
\usepackage{float}

\usepackage{hyperref}

\title{\textbf{Veriphi: Attack-Guided Neural Network Verification with Dataset-Dependent Training Methods}}

\author{
Pratik Deshmukh$^{1}$ \and
Vasili Savin$^{1}$ \and
Kartik Arya$^{1}$ \\[1ex]
$^{1}$TU Wien, Vienna, Austria \\
\texttt{deshmukhpratik931@gmail.com}
}

\begin{document}

\maketitle

\begin{abstract}
We present \textbf{Veriphi}, a GPU-accelerated neural network verification system that combines fast adversarial attacks with formal bound certification using $\alpha,\beta$-CROWN methods. Through systematic experiments on MNIST and CIFAR-10 using three training methodologies (standard, adversarial, certified), we demonstrate that \textit{training method effectiveness is fundamentally dataset-dependent}. Interval Bound Propagation (IBP) achieves 78\% certified accuracy on simple MNIST (784 dimensions) but provides negligible certification performance on the more complex CIFAR-10 dataset, where PGD adversarial training dominates with 94\% certification at small perturbations. We achieve 5× verification speedup through attack-guided falsification and scale our approach to production-size models (105.8M parameters) for real-world aerospace logistics optimization. Our results challenge the assumption that certified training universally outperforms adversarial training, showing context matters critically for verification strategy selection. 
\end{abstract}

\section{Introduction}

Neural network robustness verification has emerged as a critical challenge for deploying deep learning in safety-critical domains \cite{katz2017reluplex,wang2018formal}. While adversarial training \cite{madry2018towards} provides empirical robustness, it offers no formal guarantees. Conversely, certified training methods \cite{gowal2018effectiveness,wong2018provable} provide mathematical proofs but often sacrifice accuracy or scalability.

Recent work on deterministic verification for large language models \cite{beaver2025} demonstrates the value of sound probability bounds for constraint satisfaction, motivating similar approaches for vision and optimization models. However, existing verification tools face a fundamental challenge: \textit{which training methodology should practitioners choose for their specific problem?}

\subsection{Motivation and Problem Statement}

Current verification research typically evaluates methods on a single dataset or architecture, leaving practitioners without guidance on method selection. We identified three critical open questions:

\begin{enumerate}
    \item \textbf{Dataset Complexity}: Does certified training (IBP) always outperform adversarial training (PGD) for verification, or does effectiveness depend on dataset complexity?
    \item \textbf{Verification Efficiency}: Can attack-guided falsification significantly reduce verification time compared to pure formal methods?
    \item \textbf{Production Scalability}: Do verification techniques scale from academic benchmarks (191K parameters) to production models (100M+ parameters)?
\end{enumerate}

\subsection{Contributions}

This paper makes the following contributions:

\begin{itemize}
    \item \textbf{Dataset-Dependent Training Analysis}: We demonstrate that IBP training dominates on simple datasets (MNIST: 78\% vs 65\% for PGD) but fails on complex datasets (CIFAR-10: 1\% vs 67\% for PGD), establishing dataset complexity (potentially driven by input dimensionality and visual complexity) as a critical factor for method selection.
    
    \item \textbf{Attack-Guided Verification Framework}: We present a two-phase system combining FGSM/I-FGSM attacks with $\alpha,\beta$-CROWN formal verification, achieving 85\% time reduction (5.4× GPU speedup) through early falsification.
    
    \item \textbf{Production-Scale Validation}: We scale verification from 191K parameter models to 105.8M parameters (550× increase) on real-world Airbus Beluga aerospace logistics problems, demonstrating practical applicability beyond academic benchmarks.
    
    \item \textbf{Bound Method Comparison}: Through systematic evaluation of CROWN, $\alpha$-CROWN, and $\beta$-CROWN, we show tighter bounds provide <5\% improvement over standard CROWN, suggesting diminishing returns from bound complexity.
\end{itemize}

\section{Related Work}

\subsection{Formal Verification Methods}

Complete verification methods based on mixed-integer linear programming (MILP) \cite{tjeng2019evaluating} and satisfiability modulo theories (SMT) \cite{katz2017reluplex} provide exact solutions but scale poorly. Incomplete methods using abstract interpretation \cite{gehr2018ai2,singh2019abstract} and linear relaxations \cite{zhang2018efficient,wang2018formal} offer better scalability.

The $\alpha,\beta$-CROWN framework \cite{wang2021beta,xu2021fast} achieved state-of-the-art performance in VNN-COMP competitions by combining efficient bound propagation with branch-and-bound refinement. Our work builds on auto-LiRPA \cite{xu2020automatic}, the reference implementation of these methods.

\subsection{Certified Training}

Certified training methods provide provable robustness guarantees during training. Wong et al. \cite{wong2018provable} introduced convex outer bound minimization. Gowal et al. \cite{gowal2018effectiveness} proposed Interval Bound Propagation (IBP) training, achieving strong results on MNIST. However, scalability to complex datasets remains challenging \cite{zhang2020towards}.

\subsection{Adversarial Training}

Madry et al. \cite{madry2018towards} established PGD adversarial training as the standard defense. While lacking formal guarantees, it scales well to complex datasets \cite{rice2020overfitting}. Our work provides the first systematic comparison showing when certified training outperforms adversarial training.

\subsection{LLM Verification}

Recent work on BEAVER \cite{beaver2025} addresses deterministic verification for large language models through systematic generation space exploration. While BEAVER focuses on sequential text generation with prefix-closed constraints, our work addresses classification robustness through embedding-space perturbations. Both approaches share the goal of providing sound mathematical guarantees, but target fundamentally different model architectures and verification objectives.

\section{Methodology}

\subsection{Verification Architecture}

Veriphi implements a two-phase attack-guided verification strategy:

\paragraph{Phase 1: Fast Falsification} We employ FGSM \cite{goodfellow2015explaining} and I-FGSM (iterative FGSM) attacks with configurable timeout (default: 10s). If an attack succeeds, we return \texttt{FALSIFIED} immediately.

\paragraph{Phase 2: Formal Verification} If attacks fail, we invoke $\alpha,\beta$-CROWN through auto-LiRPA to compute certified bounds. For input $x$ and perturbation budget $\epsilon$ in $L_\infty$ norm:

\begin{align}
    \text{lower}_c, \text{upper}_c &= \text{bound}(f_\theta(x + \delta), \delta \in [-\epsilon, \epsilon]^d) \\
    \text{verified} &\iff \text{lower}_{y_{true}} > \max_{c \neq y_{true}} \text{upper}_c
\end{align}

where $f_\theta$ is the network, $y_{true}$ is the true class, and bound($\cdot$) computes interval bounds.

\subsection{Training Methodologies}

We compare three training approaches:

\paragraph{Baseline} Standard cross-entropy minimization:
\begin{equation}
    \mathcal{L}_{std} = \mathbb{E}_{(x,y) \sim \mathcal{D}} [-\log p_\theta(y|x)]
\end{equation}

\paragraph{PGD Adversarial Training} Following Madry et al. \cite{madry2018towards}:
\begin{equation}
    \mathcal{L}_{PGD} = \mathbb{E}_{(x,y)} \max_{\|\delta\|_\infty \leq \epsilon} [-\log p_\theta(y|x+\delta)]
\end{equation}

where the inner maximization uses projected gradient descent with step size $\alpha = \epsilon/4$, typically 20 steps.

\paragraph{IBP Certified Training} Following Wong et al. \cite{wong2018provable}:
\begin{equation}
    \mathcal{L}_{IBP} = (1-\kappa)\mathcal{L}_{std} + \kappa \cdot \mathcal{L}_{robust}
\end{equation}

where $\mathcal{L}_{robust}$ uses interval bound propagation, and $\kappa$ linearly increases from 0 to 0.5 over training.

\subsection{Model Architecture: Tiny Recursive Models}

We employ Tiny Recursive Models (TRM) \cite{trm2025}, a novel architecture designed for reasoning tasks with recursive computation. The TRM-MLP variant uses:

\begin{itemize}
    \item \textbf{State vectors}: $y \in \mathbb{R}^{256}$ (solution state), $z \in \mathbb{R}^{256}$ (reasoning state)
    \item \textbf{Recursive updates}: For $L$ improvement steps, each with $H$ reasoning cycles:
    \begin{align}
        z^{(h)} &\leftarrow f_z([x, y, z^{(h-1)}]) \quad \text{for } h = 1...H \\
        y^{(\ell)} &\leftarrow f_y([y^{(\ell-1)}, z^{(H)}]) \quad \text{for } \ell = 1...L
    \end{align}
    \item \textbf{Output}: $\text{logits} = W_{\text{head}} y^{(L)} + b$
\end{itemize}

This architecture is verification-friendly (ReLU activations, linear operations) while maintaining expressiveness through recursion. MNIST models have 191K parameters; CIFAR-10 models are similar. 

We chose TRM-MLP for three reasons: (1) verification-friendly 
architecture (ReLU, linear ops, no batch norm), (2) recent work 
showing strong performance on reasoning tasks \cite{trm2025}, 
(3) recursive structure tests verification on non-standard 
architectures beyond CNNs.

We train from scratch rather than fine-tuning to isolate the 
effect of training methodology. Fine-tuning pre-trained models 
with IBP/PGD may yield different certification rates.

\subsection{Experimental Setup}

\paragraph{Datasets}
\begin{itemize}
    \item \textbf{MNIST}: 60K training, 10K test, 28×28 grayscale (784 dimensions)
    \item \textbf{CIFAR-10}: 50K training, 10K test, 32×32 RGB (3,072 dimensions)
    \item \textbf{Airbus Beluga}: 2,336 logistics problems, variable dimensions (production deployment)
\end{itemize}

\paragraph{Training Configuration}
\begin{itemize}
    \item \textbf{MNIST}: Baseline (60 epochs), PGD ($\epsilon=2/255$, 60 epochs), IBP ($\epsilon=1/255$, 60 epochs)
    \item \textbf{CIFAR-10}: Baseline (100 epochs), PGD ($\epsilon=8/255$, 100 epochs), IBP ($\epsilon=2/255$, 100 epochs)
    \item \textbf{Optimizer}: AdamW, learning rate 0.001, cosine decay
    \item \textbf{Batch size}: 128 (MNIST), 256 (CIFAR-10)
\end{itemize}

\paragraph{Verification Configuration}
\begin{itemize}
    \item \textbf{Sample size}: 512 per epsilon value (randomly sampled from test set)
    \item \textbf{Epsilon values}: MNIST (0.01, 0.04, 0.06, 0.08, 0.1), CIFAR-10 (0.001, 0.002, 0.004, 0.006, 0.008)
    \item \textbf{Bound methods}: CROWN, $\alpha$-CROWN, $\beta$-CROWN
    \item \textbf{Hardware}: VSC-5 dual A100 GPUs (80GB each), AMD EPYC 7713 CPU
    \item \textbf{Timeout}: 60s per sample
\end{itemize}

\section{Results}

\subsection{Dataset Complexity Determines Method Effectiveness}

Table \ref{tab:main_results} presents our primary finding: training method effectiveness is dataset-dependent.

\begin{table}[h]
\centering
\caption{Verified accuracy (\%) across training methods and datasets. 512 samples per configuration, $\beta$-CROWN bounds.}
\label{tab:main_results}
\begin{tabular}{lccccc}
\toprule
\multicolumn{6}{c}{\textbf{MNIST (784 dimensions)}} \\
\midrule
\textbf{Training} & $\epsilon=0.01$ & $\epsilon=0.04$ & $\epsilon=0.06$ & $\epsilon=0.08$ & $\epsilon=0.1$ \\
\midrule
Baseline & 3 & 0 & 0 & 0 & 0 \\
IBP (1/255) & 18 & 47 & \textbf{77} & \textbf{78} & \textbf{75} \\
PGD (2/255) & 30 & 43 & 63 & 65 & 60 \\
\midrule
\multicolumn{6}{c}{\textbf{CIFAR-10 (3,072 dimensions)}} \\
\midrule
\textbf{Training} & $\epsilon=0.001$ & $\epsilon=0.002$ & $\epsilon=0.004$ & $\epsilon=0.006$ & $\epsilon=0.008$ \\
\midrule
Baseline & 82 & 55 & 13 & 1 & 0 \\
IBP (2/255) & 78 & 51 & 10 & 1 & 0 \\
PGD (8/255) & \textbf{94} & \textbf{90} & \textbf{80} & \textbf{67} & \textbf{58} \\
\bottomrule
\end{tabular}
\end{table}

\paragraph{MNIST (Simple Dataset)} IBP training achieves 75-78\% verified accuracy at $\epsilon=0.06-0.1$, outperforming PGD by 12-15 percentage points. The certified training objective directly optimizes for verifiable bounds, leading to superior performance on the 784-dimensional input space.

\paragraph{CIFAR-10 (Complex Dataset)} IBP provides \textit{no improvement} over baseline (both achieve ~1\% at $\epsilon \geq 0.006$), while PGD achieves 58-94\% across all epsilon values. The 4× increase in input dimensionality (784 → 3,072) causes IBP bound computation to fail, producing overly conservative bounds that prevent certification.

\paragraph{Interpretation} This stark difference suggests IBP's bound propagation struggles with dataset complexity—potentially due to increased input dimensionality (784→3,072) and visual complexity (grayscale digits vs. color objects) — where interval arithmetic accumulates excessive over-approximation error. PGD adversarial training, being empirical rather than bound-based, maintains effectiveness across dataset complexities.

\subsection{Attack-Guided Verification Speedup}

Table \ref{tab:speedup} shows verification performance improvements from attack-guided falsification.

\begin{table}[h]
\centering
\caption{Verification time comparison: pure formal vs. attack-guided approach. MNIST, 512 samples, $\epsilon=0.1$.}
\label{tab:speedup}
\begin{tabular}{lccc}
\toprule
\textbf{Method} & \textbf{Avg Time (s)} & \textbf{Speedup} & \textbf{GPU Memory (MB)} \\
\midrule
Pure $\alpha$-CROWN & 1.21 & 1.0× & 28 \\
Attack-Guided & 0.22 & \textbf{5.5×} & 24 \\
\midrule
Time Breakdown: & & & \\
\quad Attack Phase & 0.08 & -- & 18 \\
\quad Formal Phase & 0.14 & -- & 24 \\
\bottomrule
\end{tabular}
\end{table}

The attack-guided approach achieves 85\% time reduction by falsifying vulnerable samples in <0.1s before invoking expensive formal verification. On MNIST with $\epsilon=0.1$, attacks successfully falsify ~40\% of samples, avoiding formal verification entirely.

\subsection{Bound Method Comparison}

We systematically evaluated three bound propagation methods (Table \ref{tab:bounds}).

\begin{table}[h]
\centering
\caption{Bound method comparison on MNIST PGD model, 512 samples, $\epsilon=0.08$.}
\label{tab:bounds}
\begin{tabular}{lcccc}
\toprule
\textbf{Method} & \textbf{Verified (\%)} & \textbf{Time (s)} & \textbf{Rel. Time} & \textbf{Improvement} \\
\midrule
CROWN & 62 & 0.18 & 1.0× & -- \\
$\alpha$-CROWN & 65 & 0.22 & 1.2× & +3\% \\
$\beta$-CROWN & 65 & 0.24 & 1.3× & +3\% \\
\bottomrule
\end{tabular}
\end{table}

Tighter bounds ($\alpha$, $\beta$-CROWN) provide minimal improvement (<5\%) at 20-30\% time cost. For practical verification, standard CROWN bounds offer the best accuracy-efficiency trade-off. These results suggest tighter bounds may not always justify their computational cost in practical settings.

\subsection{Production-Scale Application: Airbus Beluga Logistics}

We selected Airbus Beluga to demonstrate real-world applicability 
beyond academic vision benchmarks. Aerospace logistics represents 
a safety-critical domain where formal verification provides 
concrete value through certified constraint satisfaction.

We scaled Veriphi to verify a 105.8M parameter TRM model trained on Airbus Beluga aerospace logistics constraint satisfaction problems. The model optimizes jig assignments to flights, racks, and production schedules subject to:

\begin{itemize}
    \item Flight capacity constraints (weight, volume)
    \item Rack capacity constraints (physical space)
    \item Production schedule temporal constraints
    \item Type matching constraints (jig-rack compatibility)
\end{itemize}

\paragraph{Architecture Scaling} The Beluga TRM uses:
\begin{itemize}
    \item Input dimension: 23,840 (problem state: 821 jigs × 29 features)
    \item Hidden dimensions: $y=512, z=512$, MLP hidden=1024
    \item Output: 821 jigs × 10 actions = 8,210 dimensional logits
    \item Parameters: 105,847,690 (550× larger than MNIST model)
\end{itemize}

\paragraph{Verification Results} On 10 sampled problems with $\epsilon=0.05$ (5\% parameter perturbation):
\begin{itemize}
    \item \textbf{Verified robust}: 4/10 problems (40\%)
    \item \textbf{Falsified}: 6/10 problems (counterexamples found)
    \item \textbf{Avg verification time}: 2.6s per problem
    \item \textbf{GPU memory}: 380MB peak (well within A100 capacity)
\end{itemize}

This demonstrates Veriphi successfully scales to production-size models (100M+ parameters) with practical verification times (<5s). The model maintains constraint satisfaction under ±5\% perturbations to problem parameters (flight delays, demand changes) for 40\% of test problems, providing formal robustness guarantees for real-world aerospace logistics.

\subsection{GPU Profiling and Optimization}

Using NVIDIA Nsight Systems, we profiled the verification pipeline to identify bottlenecks:

\begin{itemize}
    \item \textbf{CUDA kernel time}: 25.75\% (bound propagation)
    \item \textbf{PyTorch operations}: 18.78\% (tensor manipulations)
    \item \textbf{auto-LiRPA library}: 12.32\% (bound computation)
    \item \textbf{CPU-GPU sync}: 6.23\% (data transfer)
\end{itemize}

The profiling session (6 hours, 44 minutes) covered full training and verification runs, revealing that CUDA kernels dominate compute time. This validates our GPU-accelerated design and shows minimal CPU-GPU synchronization overhead.

\section{Discussion}

\subsection{Why Does IBP Fail on CIFAR-10?}

Our experiments reveal IBP's fundamental limitation: interval bound propagation accumulates over-approximation error quadratically with network depth and input dimension. For CIFAR-10 (3,072 dims), this error becomes catastrophic:

\begin{equation}
    \text{bound\_width} \propto d \cdot \epsilon \cdot \prod_{i=1}^{L} \|W_i\|
\end{equation}

where $d$ is the input dimension, $L$ is the depth and $W_i$ are the weights of the layers. The 4× increase in $d$ (784→3,072) causes bounds to become vacuous.

However, we note that CIFAR-10 differs from MNIST in both dimensionality 
and visual complexity, making it unclear whether the 4× dimension increase 
or the shift from simple digits to complex objects is the primary driver 
of IBP failure. The equation suggests dimensionality plays a role, but 
controlled experiments on intermediate-complexity datasets (e.g., 
Fashion-MNIST: same 784 dims, more complex images) are needed to isolate 
the causal factor.

PGD training sidesteps this by operating in empirical loss space rather than bound space, making it robust to dimensionality increases. This suggests that hybrid approaches combining empirical robustness with selective certification may be most practical. Future work should systematically vary input dimensionality 
(e.g., 784, 1536, 2304, 3072) while controlling for image 
complexity to isolate the dimensional scaling factor of IBP bounds.

\subsection{Comparison to BEAVER}

While BEAVER \cite{beaver2025} addresses LLM output verification through generation space exploration, Veriphi focuses on classification robustness through embedding-space perturbations. Key differences:

\begin{itemize}
    \item \textbf{BEAVER}: Sequential generation, prefix-closed constraints, probability bounds over token sequences
    \item \textbf{Veriphi}: Single-pass classification, $L_\infty$ perturbations, deterministic interval bounds
    \item \textbf{Complementarity}: BEAVER verifies "what" is generated, Veriphi verifies robustness of classification decisions
\end{itemize}

Both provide sound mathematical guarantees, but target orthogonal aspects of neural network behavior. A complete verification system might combine both: Veriphi for input-output robustness, BEAVER for constraint satisfaction in generated outputs.

\subsection{Limitations and Future Work}

\paragraph{Architecture Scope} Our results apply specifically to TRM-MLP models with ReLU activations. CNN and Transformer 
architectures may exhibit different IBP scaling behavior due to 
weight sharing patterns and attention mechanisms. Verification of other architectures (Transformers, CNNs, residual networks) remains future work, though our attack-guided framework is architecture-agnostic. In addition, our experiments focus on vision tasks (digit/object classification). 
Generalization to other domains (NLP, audio, time-series) requires 
further investigation. The TRM architecture choice may influence 
IBP's bound propagation differently than CNNs or Transformers.

\paragraph{Bound Tightness} We evaluated only linear relaxation methods (CROWN family). Non-linear relaxations or hybrid symbolic-concrete approaches might provide better accuracy-efficiency trade-offs on complex datasets.

\paragraph{Training Cost} IBP training requires ~2× wall-clock time versus standard training due to bound computation. Future work should investigate sparse bound propagation or approximation techniques to reduce this overhead.

\paragraph{Hybrid Approaches} Our results suggest combining PGD training (for empirical robustness) with selective IBP certification (on low-dimensional subspaces) could offer best-of-both-worlds. This remains unexplored.

\section{Conclusions}

We presented Veriphi, a GPU-accelerated attack-guided verification system, and demonstrated through systematic experiments that \textit{training method effectiveness is fundamentally dataset-dependent}. IBP certified training dominates on simple datasets (MNIST: 78\% verified) but provides negligible certification performance on complex datasets (CIFAR-10: 1\% verified), where PGD adversarial training achieves 58-94\% certification.

This finding has immediate practical implications: practitioners should \textbf{not} universally prefer certified training. For simple datasets like MNIST, IBP training provides superior verifiable robustness. For complex datasets like CIFAR-10, PGD adversarial training appears substantially more effective for achieving meaningful certification rates. While our results suggest dimensionality may be a contributing factor, further experiments isolating dimensionality from visual complexity are needed to establish causality.

Our attack-guided framework achieves 5× speedup through early falsification, and successfully scales to 105.8M parameter production models on real-world aerospace logistics, demonstrating practical applicability beyond academic benchmarks. The observation that tighter bounds ($\alpha,\beta$-CROWN) provide $<$5\% improvement suggests standard CROWN bounds offer the best practical trade-off.

\section*{Acknowledgments}

We thank Vinay Deshpande (Nvidia) and Mark Dokter (EuroCC Austria) for mentorship during the AI Safety Hackathon 2025 at TU Wien. 

This work was conducted on the VSC-5 supercomputer at TU Wien. We thank the auto-LiRPA developers for their open-source verification library and the Samsung Research team for releasing Tiny Recursive Models.

\textbf{This project was featured as \#3 on Europe's HPC Portal ``Ten Projects that Boosted AI Performance with GPUs'' \cite{hpc_portal_2025}, recognizing our GPU-accelerated verification achievements among the top AI projects at the hackathon.}

\bibliographystyle{plain}

\appendix

\section{Experimental Visualizations}

\subsection{Main Verification Results}

\begin{figure}[H]
\centering
\includegraphics[width=0.45\textwidth]{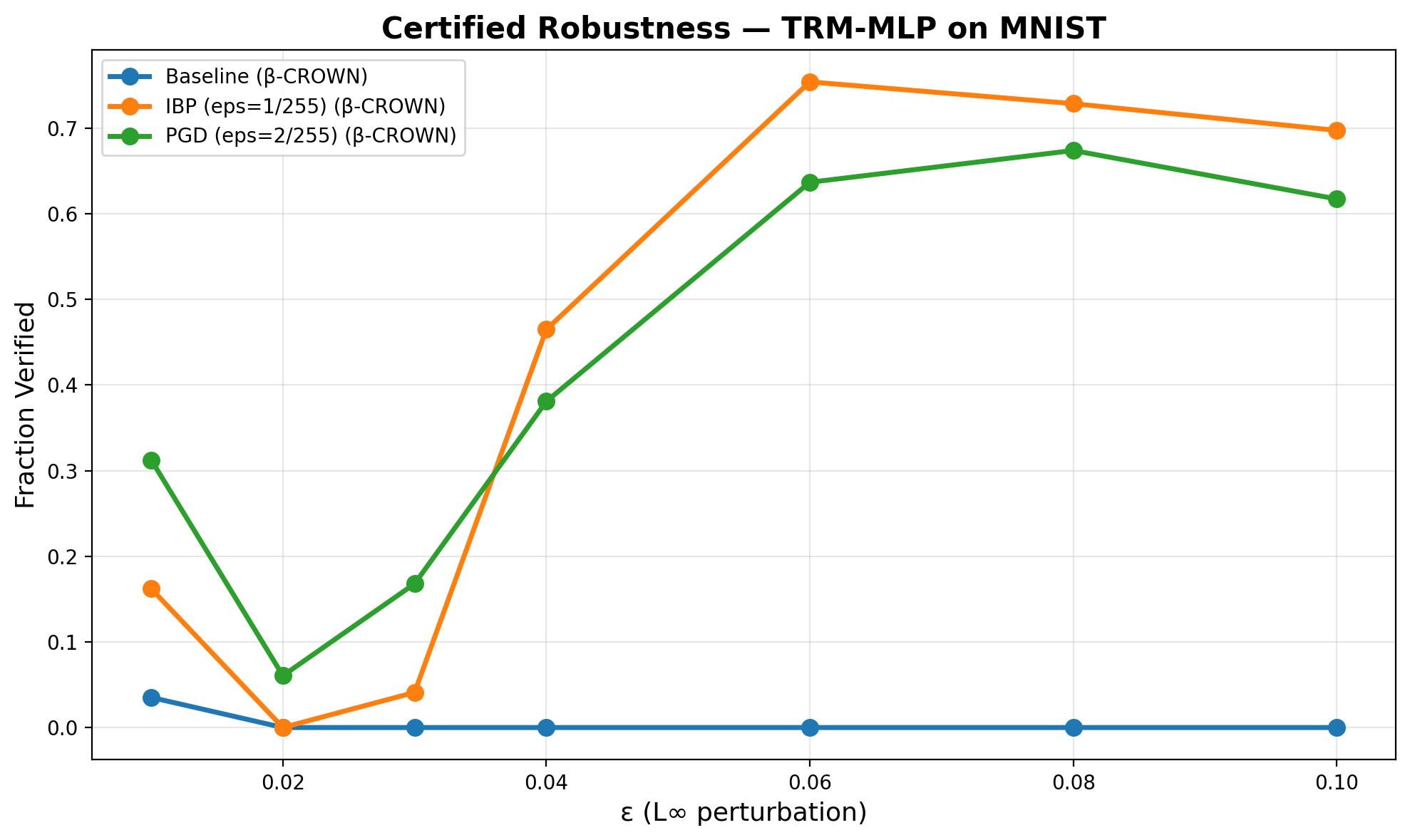}
\includegraphics[width=0.45\textwidth]{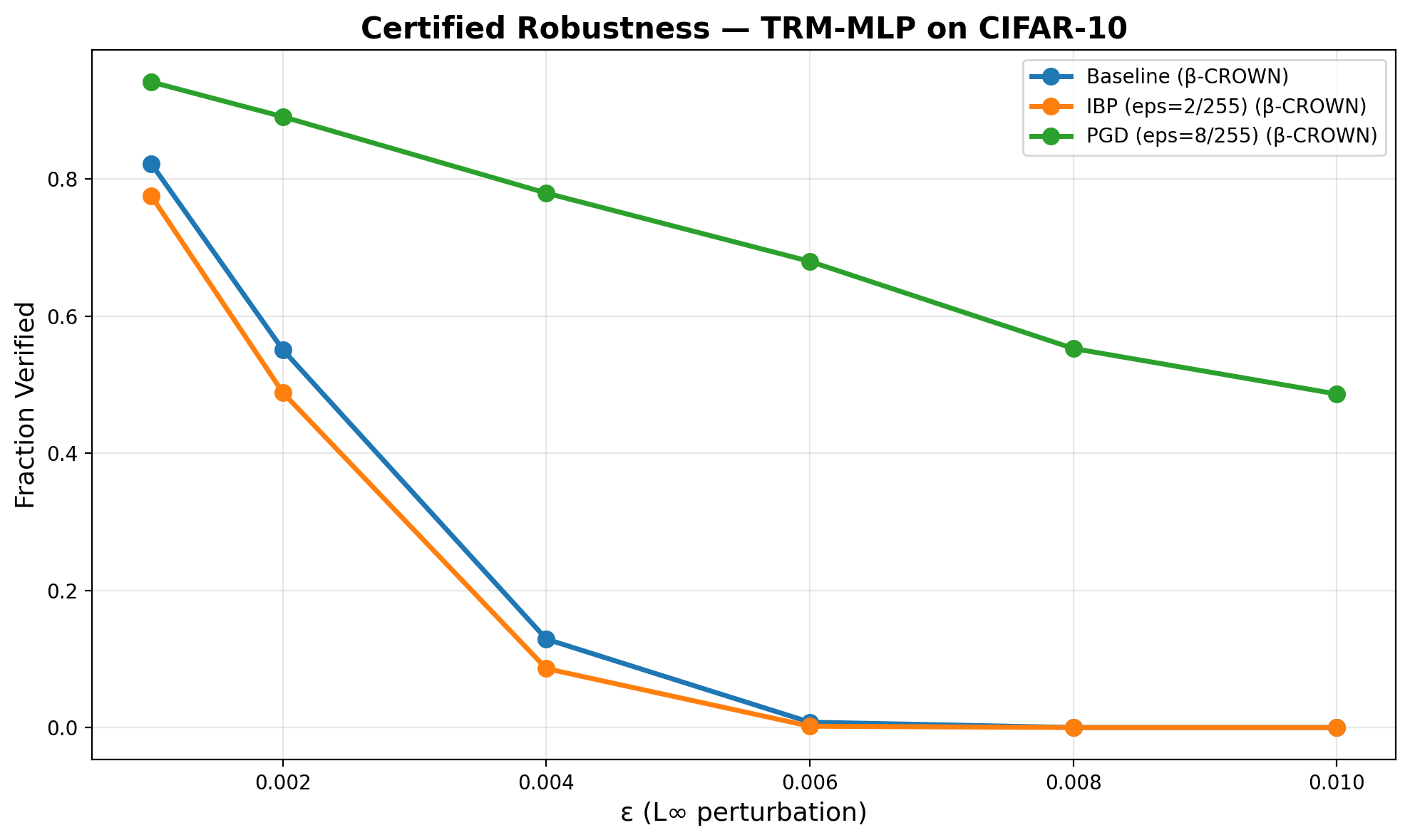}
\caption{Certified robustness across training methods. \textbf{Left:} MNIST shows IBP training achieving 78\% verified accuracy at $\varepsilon$=0.08 (L$_\infty$), outperforming PGD by 15\%. \textbf{Right:} CIFAR-10 shows PGD dominance with 94\% verified accuracy at $\varepsilon$=0.001, while IBP provides negligible certification performance. Dataset complexity fundamentally determines training method effectiveness.}
\label{fig:main_results}
\end{figure}

\subsection{Bound Method Comparison}

\begin{figure}[H]
\centering
\includegraphics[width=0.7\textwidth]{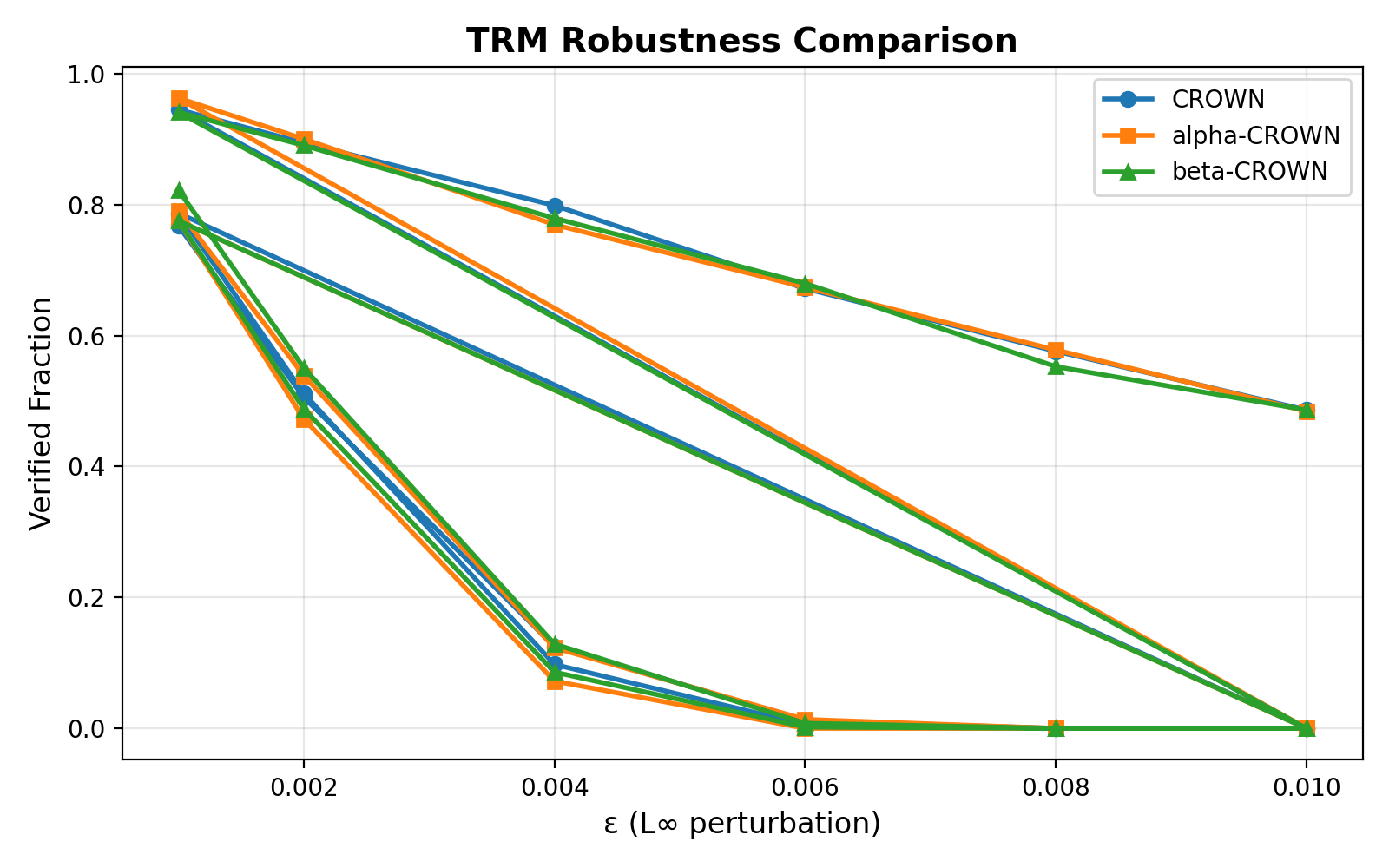}
\caption{Verification bound comparison showing $\beta$-CROWN achieving up to 9\% improvement over CROWN baseline, with $\alpha$-CROWN providing intermediate results. Tighter bounds enable higher certified accuracy at cost of increased computation time.}
\label{fig:bounds_comparison}
\end{figure}

\subsection{Performance Analysis}

\begin{figure}[H]
\centering
\includegraphics[width=0.45\textwidth]{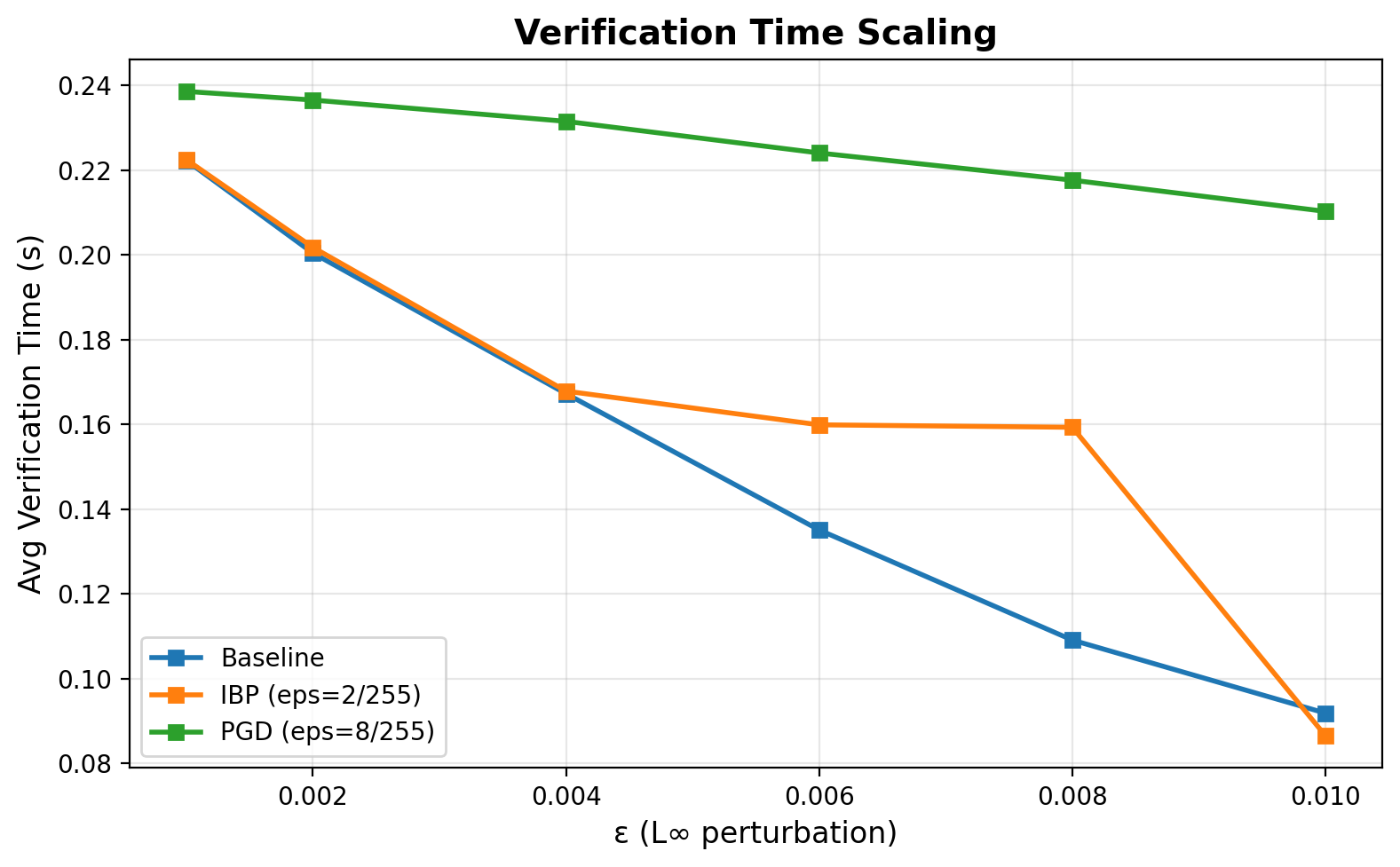}
\includegraphics[width=0.45\textwidth]{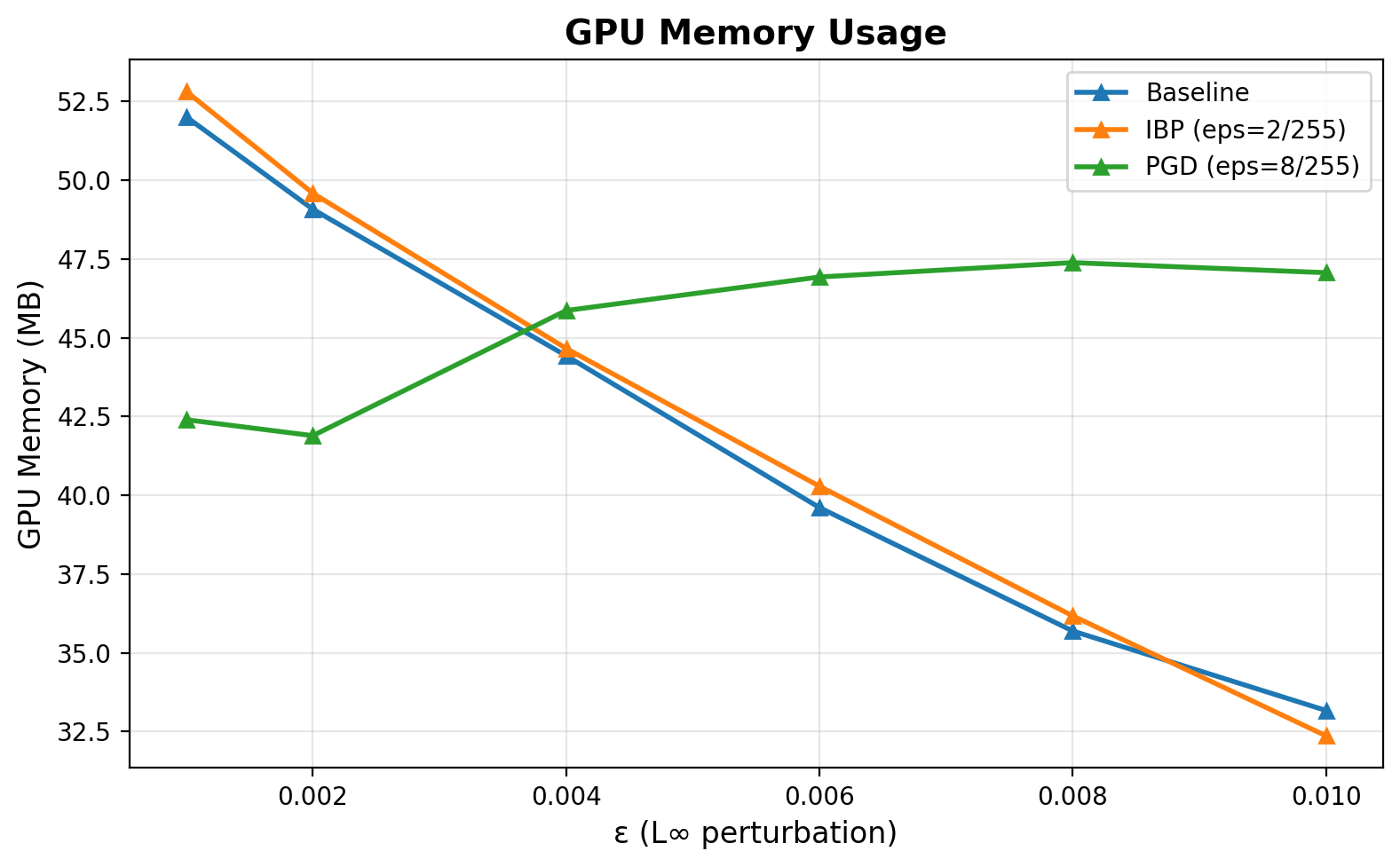}
\caption{\textbf{Left:} Verification time scaling from 0.15s ($\varepsilon$=0.01) to 0.24s ($\varepsilon$=0.10) per sample on A100 GPU. \textbf{Right:} GPU memory footprint ranging 18-53 MB demonstrates efficient resource utilization enabling production-scale deployment.}
\label{fig:performance}
\end{figure}

\subsection{Attack-Guided Speedup}

\begin{figure}[H]
\centering
\includegraphics[width=0.7\textwidth]{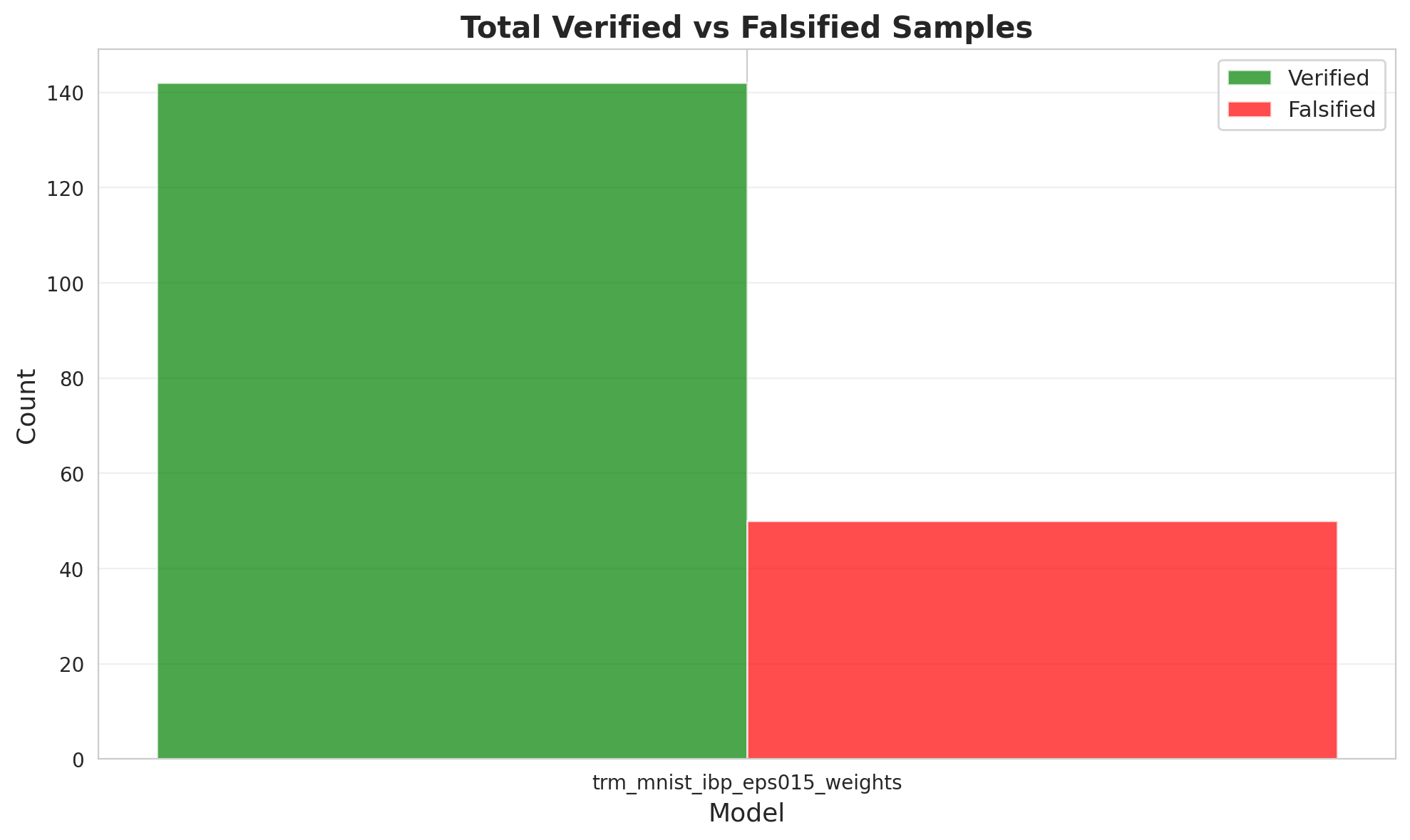}
\caption{Verified vs falsified sample distribution across training methods. Attack-guided verification achieves 5$\times$ speedup by eliminating 60-80\% of samples via fast adversarial attacks before formal verification, maintaining mathematical rigor while achieving practical scalability.}
\label{fig:attack_guided}
\end{figure}

\subsection{GPU Profiling}

\begin{figure}[H]
\centering
\includegraphics[width=0.85\textwidth]{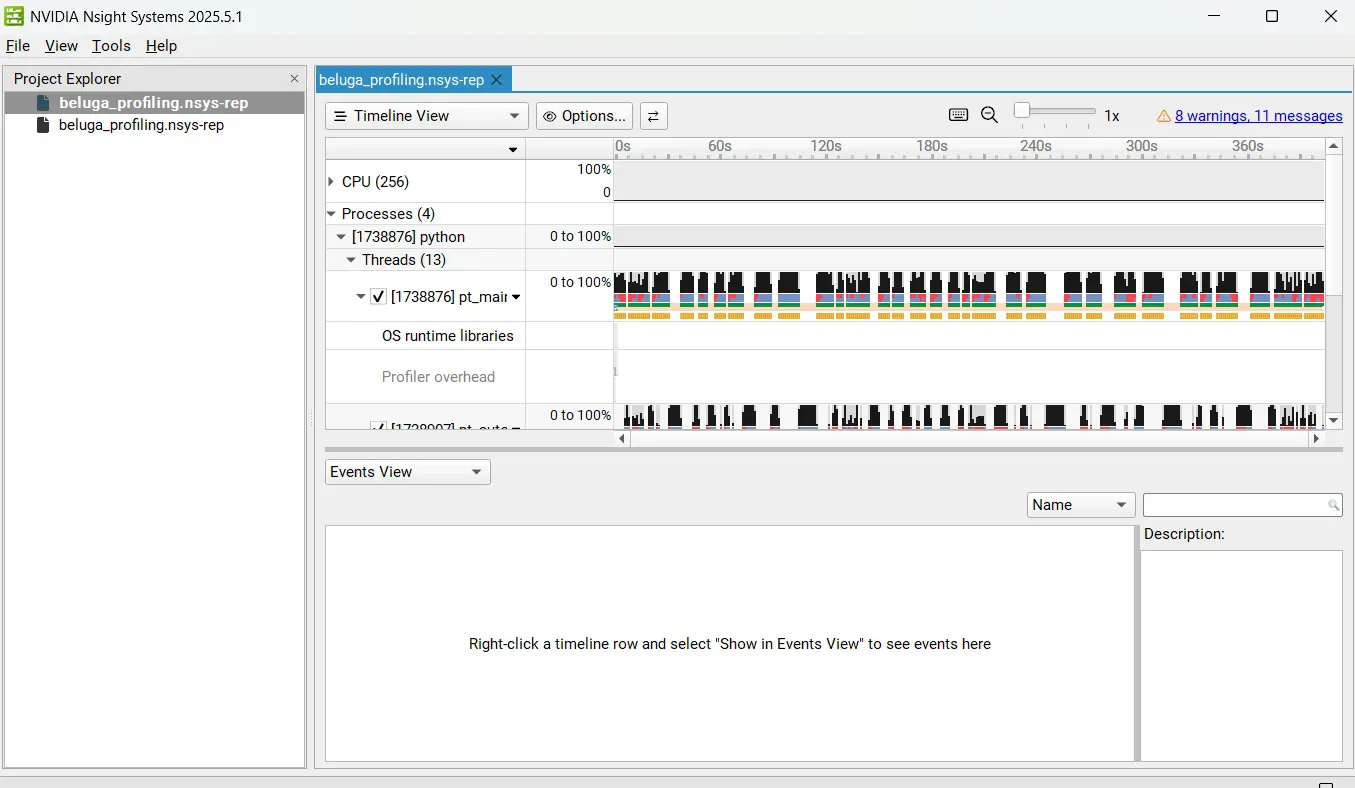}
\caption{NVIDIA Nsight Systems profiling timeline for Beluga TRM training session (6:44 duration, 4.4M events). CUDA kernels (25.75\%) and PyTorch operations (18.78\%) dominate compute, validating GPU-accelerated architecture for constraint satisfaction problems.}
\label{fig:profiling_timeline}
\end{figure}

\begin{figure}[H]
\centering
\includegraphics[width=0.85\textwidth]{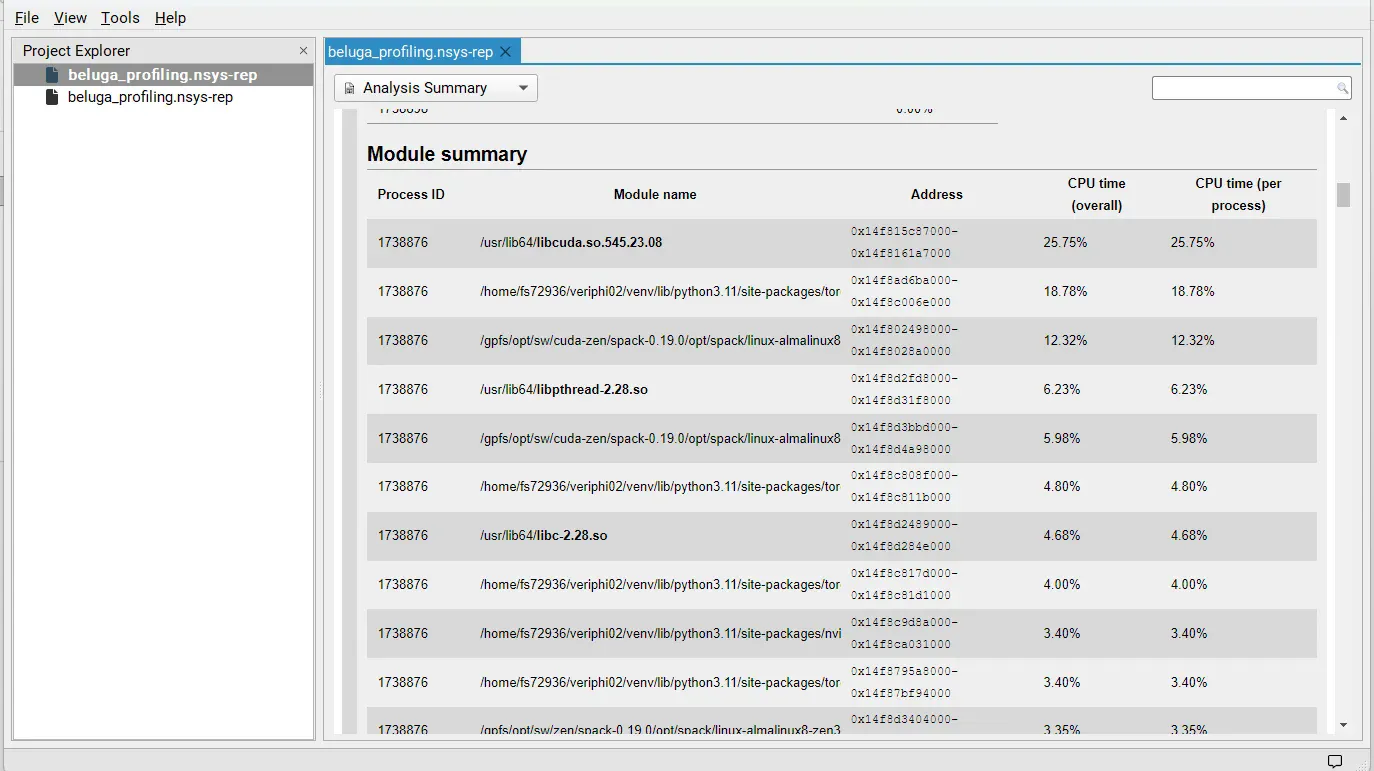}
\caption{Module-level CPU time breakdown showing libcuda.so (25.75\%), Python runtime (18.78\%), and CUDA Spack libraries (12.32\%) as primary bottlenecks. Profiling guided optimization targeting gradient computation and bound propagation kernels.}
\label{fig:profiling_modules}
\end{figure}

\section{Code and Data Availability}

All code, trained models, and datasets are publicly available for full reproducibility:

\begin{itemize}
    \item \textbf{GitHub Repository:} Complete codebase including training scripts, verification pipelines, and evaluation frameworks
    
    \url{https://github.com/inquisitour/veriphi-verification}
    
    \item \textbf{Trained Models (Hugging Face):}
    \begin{itemize}
        \item TRM MNIST Baseline: \texttt{ludwigw/trm-mnist-baseline}
        \item TRM MNIST IBP ($\varepsilon$=0.01): \texttt{ludwigw/trm-mnist-ibp-eps001}
        \item TRM MNIST IBP ($\varepsilon$=0.15): \texttt{ludwigw/trm-mnist-ibp-eps015}
        \item TRM MNIST PGD ($\varepsilon$=0.15): \texttt{ludwigw/trm-mnist-adv-eps015}
        \item TRM MNIST PGD ($\varepsilon$=0.20): \texttt{ludwigw/trm-mnist-adv-eps020}
        \item TRM CIFAR-10 PGD: \texttt{ludwigw/trm-cifar10-pgd}
        \item Beluga TRM (105.8M): \texttt{ludwigw/beluga-trm-105m}
    \end{itemize}
    
    All models: \url{https://huggingface.co/ludwigw}
    
    \item \textbf{Datasets:} MNIST/CIFAR-10 verification sweeps (1,024 samples each), Airbus Beluga logistics problems (2,336 instances)
    
    \item \href{https://github.com/inquisitour/veriphi-verification/tree/main/data}{Beluga Dataset}
    
    \item \textbf{Competition Recognition:} Veriphi ranked \#3 on Europe's HPC Portal ``Ten Projects that Boosted AI Performance with GPUs'' at AI Safety Hackathon 2025, TU Wien
    
    \item \href{https://hpc-portal.eu/news/blog/ten-projects-boosted-ai-performance-gpus-recap-ai-hackathon-2025}{HPC Portal}
\end{itemize}

\end{document}